\documentclass[letterpaper, 10pt, conference]{ieeeconf}

\IEEEoverridecommandlockouts   
\usepackage[utf8]{inputenc}
\usepackage[T1]{fontenc}

\usepackage{amsmath,amssymb}
\usepackage{graphicx}
\usepackage{booktabs}
\usepackage{array}
\usepackage{multirow}
\usepackage{caption}
\usepackage{subcaption}
\usepackage{xspace}
\usepackage{algorithm}
\usepackage{algpseudocode}

\usepackage{xcolor}
\usepackage{cite}
\usepackage{bm}
\usepackage{amsthm}
\theoremstyle{plain}

\theoremstyle{definition}

\theoremstyle{remark}

\makeatletter
\let\NAT@parse\undefined
\makeatother
\usepackage[colorlinks=true,allcolors=blue]{hyperref}
\usepackage{algorithm}
\usepackage{algpseudocode}

\newcommand{\Vset}{\mathcal{V}}
\newcommand{\Tset}{\mathcal{T}}
\newcommand{\Rset}{\mathcal{R}}
\newcommand{\Eset}{\mathcal{E}}
\newcommand{\Vrt}{\mathcal{V}^{\mathrm{rt}}}
\newcommand{\Uset}{\mathcal{U}}

\newcommand{\Qout}[2]{Q^{\mathrm{out}}_{#1,#2}}
\newcommand{\Qin}[2]{Q^{\mathrm{in}}_{#1,#2}}
\newcommand{\Hout}[2]{H^{\mathrm{out}}_{#1,#2}}
\newcommand{\Hin}[2]{H^{\mathrm{in}}_{#1,#2}}

\newcommand{\lipp}{LIPP\xspace}
\newcommand{\malipp}{MA-LIPP\xspace}
\newcommand{\seqlipp}{Seq-LIPP\xspace}

\title{
MA-LIPP: Cooperative Multi-Agent Load-Aware Informative Path Planning for Heterogeneous Robot Teams
}

\author{\authorblockN{Hojune Kim}
\authorblockA{School of Electrical and\\Computer Engineering\\
University of Southern California\\
Los Angeles, California \\
Email: hojuneki@usc.edu}
\and
\authorblockN{Guangyao Shi}
\authorblockA{School of Computer Science\\
University of Southern California\\
Los Angeles, California \\
Email: shig@usc.edu}
\and
\authorblockN{Gaurav S. Sukhatme}
\authorblockA{School of Computer Science\\
University of Southern California\\
Los Angeles, California \\
Email: gaurav@usc.edu}}

\begin{document}

\maketitle
\begin{abstract}
Field robotics missions often require physical samples to be returned to laboratories for analysis, making path planning inherently load-aware and order-dependent as accumulated samples increase payload and traversal energy costs. In single-robot Load-Aware Informative Path Planning (LIPP), this rigidly couples sensing with hauling: a solitary robot must transport every collected sample, forcing frequent depot returns that severely restrict its spatial coverage. Heterogeneous multi-robot teams can overcome this bottleneck by dividing labor—enabling high-precision samplers to collect while high-capacity carriers handle transport. However, this introduces a complex coordination challenge regarding when, where, what, and to whom handoffs should occur on top of the LIPP problem. To address this tightly coupled problem, we introduce Multi-Agent LIPP (MA-LIPP), which enables teams to cooperate through asynchronous "dead drops," allowing one robot to deposit samples for another to retrieve later without requiring synchronous rendezvous. We formulate MA-LIPP as an exact Mixed-Integer Quadratic Program (MIQP) alongside a scalable Pairwise Large-Neighborhood Search (LNS) heuristic for complex real-world applications. The heuristic matches exact optima in $95.5\%$ of certified cases and reduces weighted posterior variance by $16.1$--$19.8\%$ relative to a sequential baseline on larger instances of up to 12 robots, providing a robust framework for cooperative physical-sampling missions.
\end{abstract}

\section{Introduction}
\label{sec:intro}

Many robotic missions require robots to gather information about their environment while operating under limited energy, time, and other onboard resources. Informative path planning (IPP) provides a broad framework for such problems by jointly considering where a robot should travel and which observations it should acquire~\cite{Hollinger2014SamplingBased,dutta2025active}. Most IPP formulations, however, treat information as effectively massless, as in images or in-situ sensor measurements~\cite{JakkalaA24IPP}. In many real-world missions, including planetary exploration~\cite{Farley2020}, precision agriculture~\cite{11284891}, and other field-robotics applications~\cite{Das}, obtaining information instead requires collecting physical samples and transporting them to a depot for subsequent analysis. These settings motivate load-aware informative path planning (LIPP), which explicitly accounts for the payload accumulated through sampling~\cite{kim2026lipp}. Unlike massless observations, each physical sample remains onboard until it is unloaded, increasing the energy cost of every subsequent motion. As a result, traversal cost depends on the order in which samples are collected and transported, making LIPP fundamentally order-dependent and substantially more complex than conventional IPP.

\begin{figure}[t] 
\centering \includegraphics[width=\columnwidth]{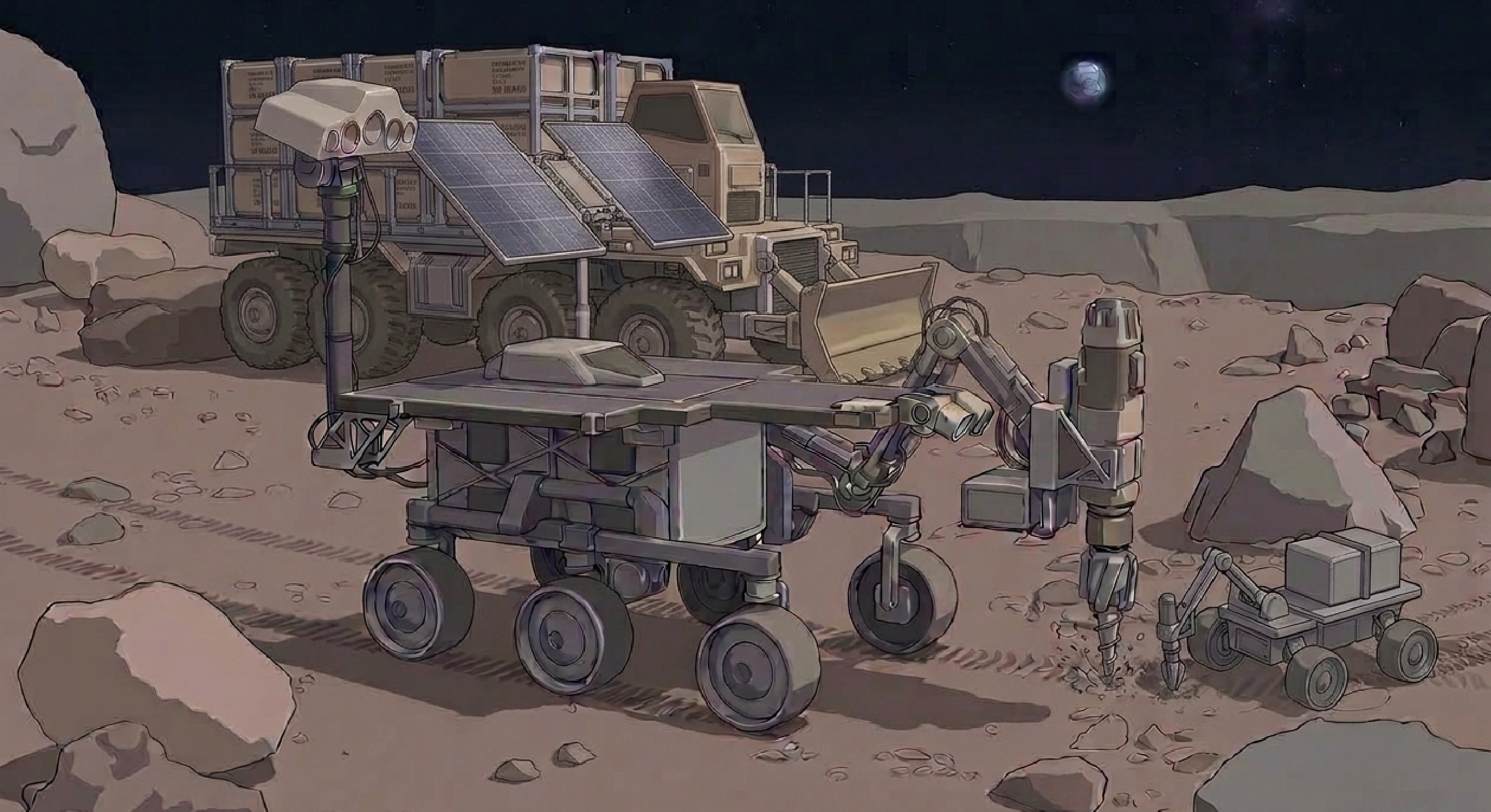} 
\caption{Planetary physical sampling with heterogeneous teams (samplers, carriers, and hybrid rovers) requires joint optimization of routing decisions, visitation order, and handoff locations/timing. MA-LIPP coordinates asynchronous sample transfers to decouple sensing from transportation, preventing payload-induced range limits and maximizing information gain.}
\label{fig:hero_fig} 
\end{figure}

As missions scale to larger environments and more demanding sampling objectives, the coverage and transport capacity of a single robot become increasingly limiting~\cite{schuster2019sampling}. Multi-robot systems offer a natural way to overcome these limitations by distributing the workload, covering spatially separated regions~\cite{singh2009efficient,JakkalaA24IPP}, and exploiting heterogeneous capabilities in sensing, mobility, and payload handling~\cite{emam2020adaptive}. This heterogeneity creates opportunities for deeper cooperation: the robot best suited to acquire a sample may not be the one best equipped to transport it. Allowing samples to be handed off between robots can therefore separate sensing from transport, enabling the team to assign each responsibility to the most capable platform.

Realizing this flexibility, however, transforms the already complex \lipp framework into a tightly coupled problem of routing, sampling, and allocation. The planner must determine not only which samples to collect in what order, but also whether, when, and to which robot each sample should be transferred. These decisions depend on the evolving state of the team, including each robot's sampling capability, sensitivity to payload, current load, and remaining route. A straightforward way to simplify the problem is to preassign sampler and carrier roles, but doing so prevents the planner from adapting as conditions change. A robot carrying samples from a teammate, for example, may later benefit from handing them off to another robot and switching back to sampling when the latter becomes better suited for transport. Effective cooperation therefore requires that sample allocation, order-dependent routing, and handoff decisions be planned jointly.

To address this challenge, we propose the \emph{Multi-Agent Load-aware Informative Path Planning (\malipp)} framework to coordinate heterogeneous teams with diverse sensing, mobility, and payload capabilities. To realize physical cooperation within this framework, our formulation models inter-robot sample transfers through asynchronous dead drops. Because sensing, transport, and routing are planned jointly, we do not prescribe fixed sampler or carrier roles; instead, the optimal division of labor emerges endogenously. 

The main contributions of this paper are as follows:
\begin{itemize}
\item We formulate the \malipp framework as an exact mixed-integer quadratic program. The formulation accommodates heterogeneity in sampling capability, mobility, payload capacity, and energy budgets; allows robot roles to change endogenously; and incorporates physically feasible inter-robot sample transfers.

\item We develop a scalable large-neighborhood search heuristic for larger \malipp instances. Across instances with certified optima, the heuristic achieves a sub-$1\%$ optimality gap in $98.2\%$ of cases, while outperforming a sequential \lipp baseline in both computation time and information gain.

\item We validate \malipp through extensive experiments across heterogeneous team settings and release a benchmark library and implementation to support future research on multi-robot physical sampling.
\end{itemize}

\section{Related Work}
\label{sec:related}

Informative path planning has been extended to multi-robot teams to coordinate where robots should collect measurements under limited sensing and motion resources. Cao et al.~\cite{cao2013multi} developed information-theoretic multi-robot path-planning methods for active sensing of spatial environmental phenomena, while Dutta et al.~\cite{dutta2025active} formulated informative path planning for Gaussian-process regression as a sparse optimization problem for resource-constrained robot teams. These methods coordinate information acquisition across multiple robots, but the measurements themselves do not become physical payloads that must subsequently be transported. Consequently, collecting a measurement does not change the cost of traversing later edges through accumulated sample mass, and there is no downstream decision of whether another robot should take responsibility for transporting what was acquired.

Kim et al.~\cite{kim2026lipp} introduced load-aware informative path planning (LIPP) for physical sampling missions, where collected samples become payload and increase the cost of subsequent motion. This makes route cost depend on sampling order. LIPP therefore captures the sampling--transportation coupling that is absent from conventional IPP, but only for a single robot. Thus, LIPP does not address how informative sampling tasks should be distributed among heterogeneous robots, how those assignments interact with multiple order-dependent routes, or when a collected sample should be transferred to another robot.

Multi-robot task allocation studies how tasks should be assigned according to robot capabilities and task requirements~\cite{gerkey2004formal,korsah2013comprehensive}, and later methods let these assignments change online: Emam et al.~\cite{emam2020adaptive} adapt specialization parameters from observed execution performance, Schillinger et al.~\cite{schillinger2021adaptive} decompose and auction formally specified tasks to heterogeneous robots, and Notomista et al.~\cite{notomista2022resilient} allocate under instantaneous energy costs. In each, the task set and its value are inputs to the allocator, and the decision is which robot executes them. In \malipp the task set is itself a decision variable: which locations are worth sampling follows from the GP posterior, so sampling value is spatially correlated, diminishing, and dependent on which robot's sensor acquires the measurement. These methods also price a task through capability matching or instantaneous energy, making assignment cost a property of the robot--task pair rather than of what the robot already carries or where the task falls in its route.

Each line thus addresses part of \malipp in isolation: multi-robot IPP selects informative measurements but treats them as massless; task allocation reallocates given tasks but prices them independently of route order and carried load; and load-coupled routing captures order dependence but with fixed, traveler-independent values and a single carrier per item. \malipp requires all three at once. Assigning a sampling task to a robot cannot be evaluated apart from the route that realizes it: collection determines when payload enters the route, how long it must be transported, and whether a later handoff would reduce total cost, while the information obtained depends on which robot acquires it. Sampling value, allocation, route order, and transport responsibility are therefore coupled within a single planning problem rather than supplied as separable task utilities and execution costs.
\section{Background}
\label{sec:background}

\subsection{LIPP Information Objective}
\label{subsec:gp}

MA-LIPP adopts and improves the information objective of load-aware informative path planning (LIPP)~\cite{kim2026lipp} literature, a single-robot framework that captures the central coupling in physical sampling: each sample improves the field estimate but also adds payload that raises the energy cost of subsequent motion.

The unknown static scalar field is modeled as a Gaussian process $f\sim\mathcal{GP}(0,k)$ with known kernel $k$, over candidate sampling nodes $\Vset$ and test locations $\Tset$. Let $k_{VV}$, $k_{TV}$, and $k_{TT}$ denote the corresponding covariance blocks, $W\succeq0$ a diagonal matrix weighting prediction errors at the test locations, and $N$ the diagonal observation-noise covariance at the sampling nodes. Since the GP posterior covariance depends on where and how precisely observations are taken, but not on their realized values, a plan's information value can be evaluated before deployment, so both LIPP and MA-LIPP are offline deterministic planning problems.

For a linear estimator with coefficient matrix $A$, where $A_{jv}$ weights the measurement at node $v$ in the estimate at test location $j$, the weighted mean-square error is
\begin{equation}
\label{eq:generic_llse}
\mathcal J(A,N)
=
\operatorname{tr}\!\left(
W\left[
A(k_{VV}+N)A^\top
-2k_{TV}A^\top
+k_{TT}
\right]
\right),
\end{equation}
Because the field and noise are jointly Gaussian, the minimum mean-square-error estimator is linear, so
\begin{equation}
\label{eq:generic_llse2}
\min_A \mathcal J(A,N)
=
\operatorname{tr}\!\left(
W\left[
k_{TT}
-k_{TV}(k_{VV}+N)^{-1}k_{VT}
\right]
\right),
\end{equation}
i.e., optimizing over $A$ recovers the weighted GP posterior variance exactly rather than approximately~\cite{dutta2025active}.

LIPP makes the noise matrix $N$ depend on the number of samples collected: averaging $c$ independent samples with noise variance $\sigma^2$ yields effective variance $\sigma^2/c$~\cite{mood1974statistics}, resulting in the noise term of the objective
\begin{equation}
\label{eq:lipp_noise_term}
\operatorname{tr}\!\left(
W\left[
A(N)A^\top
\right]
\right)
= \sum_{j\in\Tset}W_{jj}
\sum_{v\in\Vset}\sum_{c=1}^{S^{\max}}
\frac{\sigma^2}{c}A_{jvc}^2.
\end{equation}
MA-LIPP retains this estimator but, since robots may differ in sensing quality, replaces the count-based noise model with one based on the aggregate precision of the samples collected at each node (Section~\ref{subsec:sampling}).

\subsection{Inherited Load-Aware Routing Structure}
\label{subsec:lipp}

MA-LIPP also inherits LIPP's payload-dependent traversal costs and
order-dependent energy accounting. However, we replace the
Miller--Tucker--Zemlin constraints used by LIPP to eliminate disconnected
subtours with the single-commodity flow formulation presented in
Section~\ref{subsec:routing}. Flow-based connectivity provides a stronger
linear relaxation than MTZ constraints~\cite{padberg1991,oncan2009} and
also improves solution time on the single-robot LIPP instances.
\section{Problem Formulation}
\label{sec:formulation}

The \malipp formulation is organized into six parts. We first define the problem setting and information model, then the routing, handoff, and timing constraints, and finally a lexicographic objective over information quality, team energy, and mission completion time.

\subsection{Problem Setting}
\label{subsec:setting}

Let $\Rset=\{1,\dots,R\}$ denote the robot team and $\Vset$ the set of candidate sampling nodes, with $|\Vset|=n$. The physical depots $\mathcal D$ are disjoint from $\Vset$ and $\Tset$: they are neither sampled nor evaluated for prediction error. Each robot $r$ starts at $s_r$ and may unload samples and terminate its route at any depot in its nonempty authorized set $D_r\subseteq\mathcal D$, which may differ across robots. Setting $D_r=\{d\}$ for every robot gives a common single depot; with $R=1$, this recovers the usual single-robot IPP start-and-terminal structure. Robot $r$ has empty mass $m_r^0$, energy budget $B_r$, payload capacity $K_r$, per-node sampling limit $S_r^{\max}$, and measurement-noise variance $\sigma_r^2$. Each sample has unit mass $m^{\mathrm{s}}$, chosen together with the sampling limits to represent the largest mass collected at a node at sufficient resolution for payload and energy accounting. Samples are interchangeable for transportation, but their information value depends on both the sampling location and the acquiring robot's $\sigma_r^2$. We assume digitally tagged containers preserve this provenance through transfers, so routing need only track aggregate sample counts. Every collected sample must eventually be delivered to an authorized depot.

For each robot $r$, we construct a metric closure $\Eset_r$ over its given physical motion graph: each arc $(u,v)\in\Eset_r$ represents a minimum-energy physical path from $u$ to $v$. A closure arc may pass through other nodes, possibly repeatedly over a route, but only in transit, without node-level operations. Enforcing a simple route over the closure therefore does not exclude physical routes that revisit intermediate nodes. If robot $r$ carries $Q$ samples along a physical path $P$, its traversal energy is $\bigl(\sum_{e\in P}\alpha_e^r\bigr)(m_r^0+m^{\mathrm{s}}Q)$. Since the payload multiplier is common to every edge, the minimum-$\alpha$ path is payload-independent, and the closure can be constructed offline.

\subsection{Heterogeneous Sensing and Information Model}
\label{subsec:sampling}

Let $a_v^r\in\mathbb Z_{\ge0}$ denote the number of samples robot $r$ acquires at node $v$, and let $y_v^r\in\{0,1\}$ indicate that $r$ visits $v$. Sampling occurs only at visited candidate nodes $v\in\Vset$, with $a_v^r=0$ at all other routing nodes, including depots and terminal copies:
\begin{equation}
0\le a_v^r\le S_r^{\max}y_v^r.
\label{eq:sampling_visit}
\end{equation}
MA-LIPP generalizes LIPP's count-based noise model $\sigma^2/c$  of \eqref{eq:lipp_noise_term} via inverse-variance weighting~\cite{taylor1997error}. Under independent Gaussian noise, samples contribute additively to the aggregate precision $p_v$ at node $v$, and the effective observation-noise variance at a sampled node is its inverse:

\begin{align}
p_v&=\sum_{r\in\Rset}\frac{a_v^r}{\sigma_r^2},
& \rho_v&=\frac{1}{p_v}
\label{eq:precision}
\end{align}

Each robot's sample count is thus weighted by its measurement precision. Setting $N_{vv}=\rho_v$ in \eqref{eq:generic_llse} incorporates this team-level model; with a common variance $\sigma^2$, it reduces to LIPP's $\rho_v=\sigma^2/\sum_r a_v^r$.

To preserve the MIQP structure, let $p_{v1},\dots,p_{vL_v}$ denote the distinct nonzero precision levels attainable under \eqref{eq:sampling_visit}, with $\rho_{vi}=1/p_{vi}$. Binary $z_{vi}$ indicates that the team realizes level $p_{vi}$ at $v$, and the estimator coefficients $A_{jvi}$ are active only for the selected level:
\begin{align}
p_v&=\sum_{i=1}^{L_v}p_{vi}z_{vi},
& \sum_{i=1}^{L_v}z_{vi}&\le1,
\label{eq:precision_select}\\
A_{jv}&=\sum_{i=1}^{L_v}A_{jvi},
& |A_{jvi}|&\le A_j^{\max}z_{vi},
\label{eq:A_activation}
\end{align}
where $A_j^{\max}$ is a valid coefficient bound, so all coefficients vanish at unsampled nodes. Substituting into \eqref{eq:generic_llse} gives the MA-LIPP information objective
\begin{equation}
\label{eq:llse_objective}
\begin{aligned}
J(A,z)
={}&\sum_{j\in\Tset}W_{jj}\Bigg[
\sum_{u,v\in\Vset}k_{VV}(u,v)A_{ju}A_{jv}\\
&\quad+\sum_{v\in\Vset}\sum_{i=1}^{L_v}\rho_{vi}A_{jvi}^{2}\\
&\quad-2\sum_{v\in\Vset}k_{TV}(j,v)A_{jv}
+k_{TT}(j,j)\Bigg],
\end{aligned}
\end{equation}
which depends on $z$ through \eqref{eq:A_activation}. It is convex quadratic in the estimator coefficients, and minimizing over them for a fixed sampling plan recovers the exact weighted GP posterior variance.

\subsection{Robot Routing Model}
\label{subsec:routing}

Let $x_{uv}^r\in\{0,1\}$ indicate that robot $r$ traverses arc $(u,v)$. Each route starts at $s_r$, conserves arcs at intermediate nodes, and ends at exactly one terminal copy node. To allow intermediate unloading while the terminal node has no outgoing arcs, each depot $d\in D_r$ receives a terminal copy $\tilde d_r$ with the same incoming arcs and no outgoing arcs. Denoting the copies by $\tilde\Vset_r$, the routing-node and unloading-node sets are
\begin{equation}
{\Vrt}_r=\Vset\cup\mathcal D\cup\{s_r\}\cup\tilde\Vset_r,
\qquad
\Uset_r=D_r\cup\tilde\Vset_r,
\end{equation}
so a robot can unload at a physical depot, continue its route, and later terminate through a depot's copy. The routing constraints are
\begin{align}
\sum_w x_{s_rw}^r &= 1,
& \sum_u x_{us_r}^r &= 0,
\label{eq:start}\\
y_{s_r}^r &= 1,
& \sum_u x_{uv}^r &= y_v^r,
\label{eq:visit_terminal}\\
\sum_{\tilde d\in\tilde\Vset_r} y_{\tilde d}^r &= 1,
\label{eq:terminal_select}\\
\sum_w x_{vw}^r &= \sum_u x_{uv}^r,
\label{eq:balance}
\end{align}
where the second equality in \eqref{eq:visit_terminal} applies to $v\in\Vrt_r\setminus\{s_r\}$, and \eqref{eq:balance} to intermediate nodes $v\in\Vrt_r\setminus(\{s_r\}\cup\tilde\Vset_r)$.

To eliminate disconnected subtours, we replace LIPP's MTZ constraints with a single-commodity auxiliary flow $\phi_{uv}^r\ge0$. The start supplies one unit for each other visited node, and each such node consumes one unit:
\begin{align}
x_{uv}^r&\le\phi_{uv}^r\le N_r x_{uv}^r,
\label{eq:flow}\\
\sum_u\phi_{uv}^r-\sum_w\phi_{vw}^r&=y_v^r,
\label{eq:flow_balance}\\
\sum_w\phi_{s_rw}^r-\sum_u\phi_{us_r}^r
&=\sum_{v\in\Vrt_r\setminus\{s_r\}}y_v^r,
\label{eq:flow_source}
\end{align}
where \eqref{eq:flow_balance} applies to $v\in\Vrt_r\setminus\{s_r\}$, and $N_r=|\Vset\cup\mathcal D\cup\{s_r\}|$ bounds the number of visited nodes other than the start, since exactly one terminal copy is selected. Because the flow uses only selected arcs, every visited node must be connected to the start.

\subsection{Payload, Custody, and Handoff}
\label{subsec:custody}

Unlike the auxiliary flow, $Q_{uv}^r\in\mathbb Z_{\ge0}$ is the physical number of samples robot $r$ carries along arc $(u,v)$. Payload capacity and the energy budget require
\begin{align}
Q_{uv}^r&\le K_rx_{uv}^r,
\label{eq:payload}\\
\sum_{(u,v)\in\Eset_r}\alpha_{uv}^r
\bigl(m_r^0x_{uv}^r+m^{\mathrm{s}}Q_{uv}^r\bigr)
&\le B_r.
\label{eq:energy_budget}
\end{align}
Collecting, receiving, or unloading samples thus changes the energy cost of every subsequent arc, preserving the order dependence of \lipp.

Robots may transfer samples through asynchronous dead drops: a sender deposits samples at a shared physical node for a receiver to collect later. Let $H_v^{r\to r'}\in\mathbb Z_{\ge0}$ denote the transferred count and $\delta_v^{r\to r'}\in\{0,1\}$ indicate a positive transfer; handoffs are defined only at physical nodes shared by both robots' routing sets, never at terminal copies. Let $g_v^r\in\mathbb Z_{\ge0}$ denote unloading at $v\in\Uset_r$, with $g_v^r=0$ elsewhere. With $\Qin{v}{r}$, $\Qout{v}{r}$ the incoming and outgoing carried counts and $\Hin{v}{r}$, $\Hout{v}{r}$ the corresponding handoff totals, custody is conserved at every routing node:
\begin{equation}
\Qin{v}{r}+a_v^r+\Hin{v}{r}
=\Qout{v}{r}+\Hout{v}{r}+g_v^r.
\label{eq:conservation}
\end{equation}
Summing over robots and nodes cancels carried and transferred flows, so every collected sample is ultimately unloaded. Transport needs only aggregate counts; $a_v^r$ retains the sensing provenance used by the information model.

The next constraints limit unloading and outgoing deposits to the samples available from arrival and collection, keep the remaining payload within capacity after unloading, and bound the total outgoing deposit:
\begin{align}
g_v^r+\Hout{v}{r}&\le\Qin{v}{r}+a_v^r,
\label{eq:source_feasibility}\\
\Qin{v}{r}+a_v^r&\le K_r+g_v^r,
\label{eq:cap_feasibility}\\
\Hout{v}{r}&\le K_r.
\label{eq:handoff_capacity}
\end{align}
These guarantee that a feasible order of operations exists within each visit without explicitly scheduling it. In particular, \eqref{eq:conservation} and \eqref{eq:payload} imply that incoming handoffs fit within the space left after collection, unloading, and outgoing transfers:
\begin{equation}
\Hin{v}{r}\le K_r-
\bigl(\Qin{v}{r}+a_v^r-g_v^r-\Hout{v}{r}\bigr).
\label{eq:receiver_capacity}
\end{equation}

Positive transfers are activated and require both robots to visit the transfer node:
\begin{align}
\delta_v^{r\to r'}\le H_v^{r\to r'}
&\le\overline H^{r\to r'}\delta_v^{r\to r'},
\label{eq:delta_amount}\\
\delta_v^{r\to r'}&\le y_v^r,
&
\delta_v^{r\to r'}&\le y_v^{r'},
\label{eq:delta_visits}
\end{align}
where $\overline H^{r\to r'}=\min\{K_r,K_{r'}\}$ is a constant activation bound; the tighter sender-inventory and receiver-capacity limits follow from \eqref{eq:source_feasibility}--\eqref{eq:receiver_capacity}. Finally, unloading requires a visit to an authorized depot:
\begin{equation}
g_v^r\le K_r y_v^r,
\qquad v\in\Uset_r,
\label{eq:unload_bound}
\end{equation}
with the amount further limited by \eqref{eq:source_feasibility} to the samples brought into the visit, minus those deposited for other robots.

\subsection{Asynchronous Dead-Drop Timing}
\label{subsec:timing}

Let $t_v^r$ denote robot $r$'s scheduled departure time from node $v$; at a selected terminal copy, $t_{\tilde d}^r$ is its mission completion time. Each visit reserves a service window $[t_v^r-\beta,t_v^r]$, where $\beta\ge0$ conservatively bounds the time for collection, unloading, and handoff operations across all robots and nodes, given their service rates and permitted sample volumes. Travel and service times satisfy
\begin{align} t_{s_r}^r&\ge\beta, \label{eq:start_time}\\ t_v^r&\ge t_u^r+\tau_{uv}^r+\beta-\mathcal M_{uv}^{t,r}(1-x_{uv}^r). \label{eq:time_propagation} \end{align}
An optional clearance margin $\Delta^H\ge0$ further separates the sender's departure from the receiver's service window at a handoff node ($\Delta^H=0$ if none is needed). Active handoffs satisfy
\begin{align} t_v^{r'}&\ge t_v^{r}+\beta+\Delta^H-\mathcal M_v^{H,r\to r'}(1-\delta_v^{r\to r'}), \label{eq:deaddrop_sync}\\ \delta_v^{r\to r'}+\delta_v^{r'\to r}&\le1, \qquad r<r'. \label{eq:no_mutual} \end{align}
For an active handoff, \eqref{eq:deaddrop_sync} gives $t_v^{r'}-\beta\ge t_v^r+\Delta^H$: the receiver begins service only after the sender departs and the clearance margin elapses, so no simultaneous rendezvous is required. Departure times satisfy $\underline t_v^ry_v^r\le t_v^r\le\overline t_v^ry_v^r$, which zeroes times at unvisited nodes, and the constants $\mathcal M_{uv}^{t,r}$ and $\mathcal M_v^{H,r\to r'}$ deactivate timing constraints when their indicators are zero.

\subsection{Lexicographic Objective}
\label{subsec:objective}

Total team energy and mission completion time are defined by
\begin{align}
E^{\mathrm{tot}}&=\sum_{r\in\Rset}\sum_{(u,v)\in\Eset_r}\alpha_{uv}^r\bigl(m_r^0x_{uv}^r+m^{\mathrm{s}}Q_{uv}^r\bigr),
\label{eq:total_energy}\\
T^{\mathrm{end}}&\ge t_{\tilde d}^r,
\qquad r\in\Rset,\;\tilde d\in\tilde\Vset_r.
\label{eq:makespan}
\end{align}
MA-LIPP uses the lexicographic objective
\begin{equation}
\operatorname{lexmin}\bigl(J(A,z),E^{\mathrm{tot}},T^{\mathrm{end}}\bigr),
\label{eq:lex_objective}
\end{equation}
which first selects the most informative feasible team sampling plan, then minimizes the energy required to realize it, and finally minimizes mission completion time. No sampler or carrier roles are imposed beforehand; they emerge from heterogeneous sensing value, payload-dependent transport cost, and handoff opportunities. The decision variables are
\begin{equation}
\begin{aligned}
x_{uv}^r,\,y_v^r,\,z_{vi},\,\delta_v^{r\to r'}&\in\{0,1\},
&\phi_{uv}^r,\,t_v^r,\,T^{\mathrm{end}}&\in\mathbb R_{\ge0},\\
a_v^r,\,Q_{uv}^r,\,H_v^{r\to r'},\,g_v^r&\in\mathbb Z_{\ge0},
&A_{jv},\,A_{jvi}&\in\mathbb R.
\end{aligned}
\label{eq:domains}
\end{equation}
The information stage is a mixed-integer convex quadratic program; once the selected precision levels are fixed, the remaining stages are mixed-integer linear programs. The formulation contains $O(RN^2)$ routing, connectivity-flow, and payload variables and $O(R^2N)$ transfer variables, where $N=\max_r N_r$, plus $\sum_v L_v$ precision-selector binaries, where $L_v$ depends on the number of distinct sensor classes rather than directly on the number of robots.

Equations \eqref{eq:sampling_visit}--\eqref{eq:domains} define the MA-LIPP model. The implementation further tightens big-$M$ constants, removes robot--node assignments that cannot appear in any feasible mission, and adds redundant valid inequalities that strengthen the relaxation without altering the feasible set.

\begin{figure*}[!t]
\centering
\includegraphics[width=\textwidth]{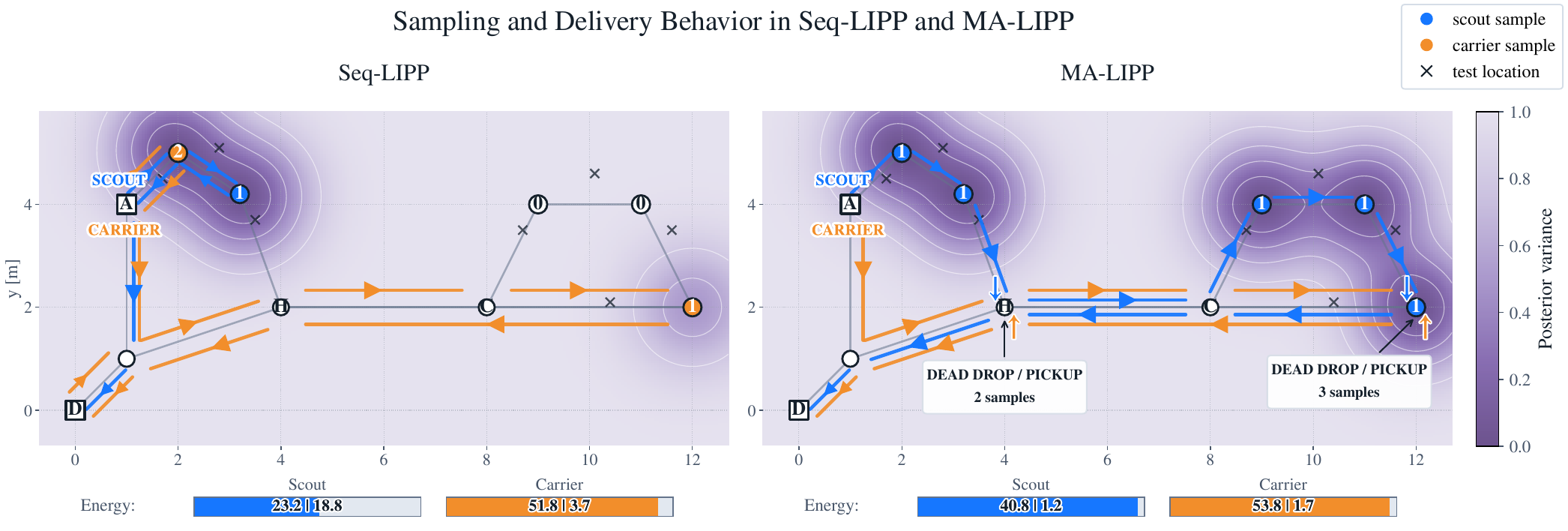}
\caption{Under Seq-LIPP, the scout cannot reach the right region after sampling the left because it must carry its accumulated payload. MA-LIPP enables the scout to drop its left-region samples at $H$, cross the corridor empty-handed, sample the right region, drop there, and return to the depot empty-handed. A more capable carrier then retrieves and consolidates all samples for final delivery, decoupling information acquisition from transport.}

\label{fig:behavior_comparison}
\end{figure*}
\section{Heuristic: Large-Neighborhood Search}
\label{sec:heuristic}

The exact MIQP yields optimal plans for small teams, but its computational cost scales poorly for large missions. We therefore introduce a pairwise large-neighborhood search (LNS) heuristic (Algorithm~\ref{alg:pairwise_lns}) that iteratively improves the team plan by letting only two robots reroute within each neighborhood, while retaining team-wide coordination of sampling and sample transfers. Because it relies on time-limited subproblem solves, it provides no global optimality certificate. However, as evaluated in Section~\ref{subsec:exp_scalability}, it reliably recovers known optima on small instances and offers a highly favorable trade-off between information gain and computation time relative to \seqlipp at larger scales. It proceeds in three phases.

\begin{algorithm}[H]
\caption{Pairwise LNS for MA-LIPP}
\label{alg:pairwise_lns}
\begin{algorithmic}[1]
\State $\mathcal{S} \gets$ Initialize via \seqlipp (fallback: direct-depot routes, then unrestricted solve)
\If{$\mathcal{S} = \emptyset$} \Return no feasible plan found \EndIf
\While{time and sweep budgets remain}
    \State $\mathrm{improved} \gets \mathrm{false}$; rank robot pairs by slack, shared nodes, and past failures
    \For{each pair $\{r,r'\}$ in ranked order}
        \State $\mathcal{S}' \gets$ Solve with only $r,r'$ rerouted, warm-started with $\mathcal{S}$, under time limit
        \State \textbf{if} $J(\mathcal{S}') < J(\mathcal{S}) - \epsilon$ \textbf{then} $\mathcal{S} \gets \mathcal{S}'$; $\mathrm{improved} \gets \mathrm{true}$
        \State Update the pair's failure penalty
    \EndFor
    \If{\textbf{not} $\mathrm{improved}$} \textbf{break} \EndIf
\EndWhile
\State \Return $\mathcal{S}$ after refining energy and mission completion time with remaining time
\end{algorithmic}
\end{algorithm}

\paragraph{Initialization} Routes are initialized with \seqlipp, a natural composition of single-robot LIPP~\cite{kim2026lipp} and sequential allocation~\cite{singh2009efficient} enhanced with mid-route unloading, in which each robot plans against the posterior conditioned on the robots committed before it. To prevent a fixed, sensing-blind order (e.g., one based purely on energy efficiency) from prematurely committing a coarse sensor to a high-value node, the order is chosen adaptively: at each step, every uncommitted robot is evaluated against the current posterior, and the one yielding the greatest team objective reduction is committed. If this fails to produce a feasible schedule, we fall back to direct-depot routes and, if those also fail, to a brief unrestricted solve. With the resulting routes frozen, \malipp jointly optimizes sampling, payload transport, and timing to form the initial incumbent.

\paragraph{Pairwise improvement} For each robot pair, we release the pair's route decisions and fix all other routes to the incumbent. Sampling, payload, handoff, unloading, and timing decisions remain free for the entire team, so robots with fixed routes can still participate in revised transfers. Each subproblem is warm-started with the incumbent, and a candidate is accepted only if it reduces the information objective by more than a tolerance $\epsilon$. Pairs are ranked by energy-slack asymmetry and shared reachable nodes, with a penalty for unsuccessful attempts. Each pair receives a fixed minimum time allowance, and the search stops when its time or sweep budget is exhausted or a full sweep yields no accepted improvement. Since only strictly improving moves are accepted, the final solution is guaranteed to perform at least as well as the initialization in the information objective.

\paragraph{Final refinement} With the incumbent's routes and precision levels fixed, the remaining time is used to minimize team energy and mission completion time $T^{\mathrm{end}}$ over custody decisions alone: which robot acquires each sample, where transfers occur, and where samples are unloaded. Both steps are small MILPs, which keeps refinement inexpensive, and neither alters the incumbent's information value.

\section{Experiments}
\label{sec:experiments}

\begin{figure}[t] 
\centering \includegraphics[width=\columnwidth]{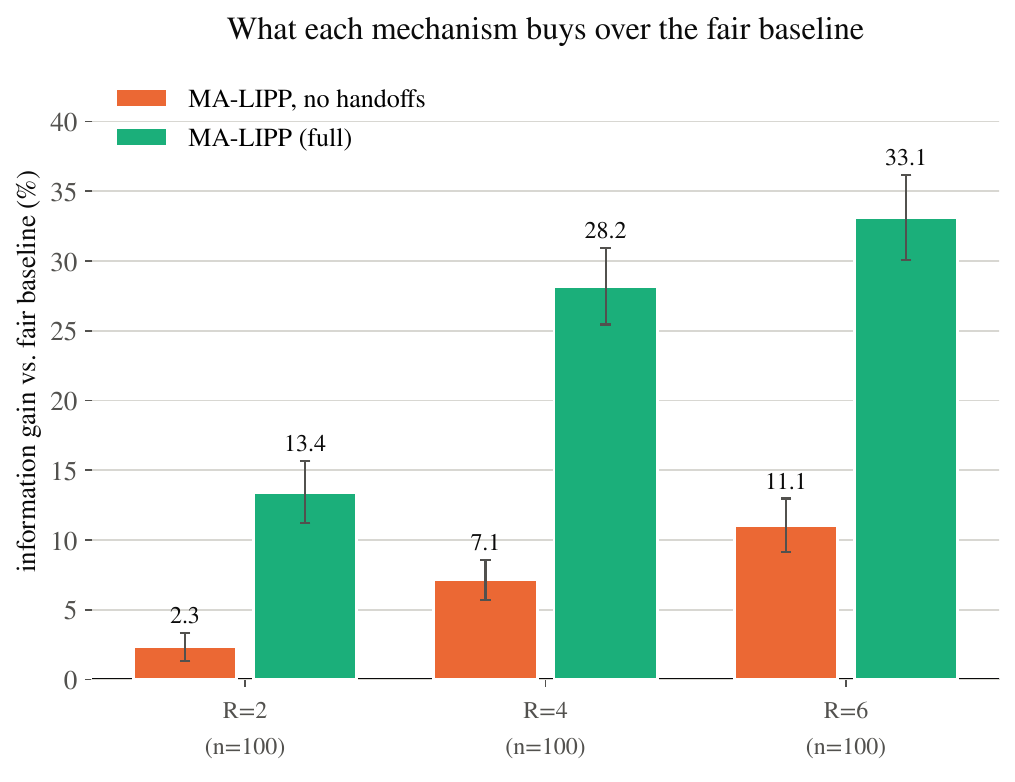} 
\caption{Information gain over \seqlipp for joint planning without handoffs versus full \malipp across three team sizes. The gap between bars highlights the incremental value of handoff. Error bars denote $95\%$ confidence intervals across $100$ instances per bin. \emph{Takeaway:} Both joint planning and handoffs consistently outperform sequential routing, with physical transfers contributing roughly three times more value than joint routing alone—an advantage that widens as fleet size increases.}\label{fig:exp1} 
\end{figure}

In this section, we evaluate three claims: (1) both the joint planning and dead-drop handoff features of \malipp improve information collection over sequential planning; (2) the value of cooperation depends on sensing heterogeneity and sample transport cost; and (3) the pairwise LNS heuristic closes the optimality gap against exact \malipp on verifiable instances and outperforms baselines in both information gain and computation time at larger scales. To evaluate these claims, we compare \seqlipp against \malipp (with/without handoffs) across difficulty levels (Fig.~\ref{fig:exp1}) and assess the impact of sensing heterogeneity and sample mass on cooperative gain (Fig.~\ref{fig:exp2}). Lastly, we measure the LNS heuristic's optimality gap on small instances and benchmark its scalability against \seqlipp on larger problems.

Directly applying off-the-shelf task allocation or multi-robot IPP methods would require either pre-computing information objective value to all possible cases or making ad-hoc repairs to energy constraints (Sec.~\ref{sec:related}). We therefore evaluate \seqlipp and a no-handoff ablation as controlled baselines within a unified formulation. As detailed in Sec.~\ref{sec:heuristic}, our baseline \seqlipp incorporates both adaptive ordering and mid-route unloading, granting it identical individual sampling and delivery capabilities to \malipp. However, inter-robot handoffs are structurally impossible under sequential planning, as they demand coupled, simultaneous multi-agent spatio-temporal synchronization. As illustrated in Fig.~\ref{fig:behavior_comparison}, this restriction forces \seqlipp agents to self-deliver their accumulated payloads, whereas \malipp coordinates handoffs so that high-quality sensing robots can delegate delivery tasks across long distances to efficient carriers and focus on sampling.
 
We report information gain as the percentage reduction in weighted posterior variance relative to the sequential baseline on the same instance, with larger values indicating greater improvement: 
$ G=100\frac{J_{\mathrm{seq}}-J_{\mathrm{method}}}{J_{\mathrm{seq}}}\;(\%). \label{eq:experimental_gain} $

Unless otherwise stated, instances use random directed graphs in a $10\times10$ workspace, edge density $0.30$, one-way-edge fraction $0.05$, and identical robot speeds. An RBF kernel with length scale $2.0$ models the field; the identity-weighted posterior variance is used $W_{jj}=1$. All nodes except starts and depots permit sampling. Sample mass is $m^{\mathrm{s}}=\lambda=1$, and $\beta=\Delta^H=0$ isolates routing and transport costs from service delays. The default solver time limit is $900\,\mathrm{s}$ per instance, with a relative MIP-gap target of $10^{-4}$ for MIP solves. Error bars indicate $95\%$ confidence intervals of the mean. We assign heterogeneous capabilities by giving higher-precision sensors to platforms with lower payload capacity and reach capability.

\subsection{Joint Planning and Dead Drops}
\label{subsec:exp_mechanisms}

In this experiment, we compare a no-handoff baseline—which optimizes routes jointly but suppresses dead drops— and a full \malipp against \seqlipp across different problem complexities. For increasing difficulty of the problem, we increase team sizes $R \in \{2, 4, 6\}$  while scaling the workspace so that we maintain a fixed workload of exactly three informative nodes per robot. Energy budgets are scaled proportionally to enforce consistently tight range constraints and maintain a uniform baseline difficulty across all sizes. We set a $20\times$ disparity in sensor noise across the fleet. Finally, solver time limits were set sufficiently high to ensure all runs terminated naturally without truncating the search.

Figure~\ref{fig:exp1} shows that both mechanisms consistently improve upon sequential planning, with performance margins widening as the fleet grows. Joint planning alone achieves information gains over \seqlipp of $2.33\%$, $7.14\%$, and $11.10\%$ at $R=2$, $4$, and $6$. Permitting dead drops contributes roughly three times more additional value, driving total \malipp gains to $13.44\%$, $28.19\%$, and $33.10\%$. In every instance, full \malipp strictly matched or outperformed its no-handoff counterpart. This expanding advantage stems directly from increased physical cooperation, with average dead drops per mission rising from $1.61$ at $R=2$ to $10.06$ at $R=6$. Because the per-robot workload remains constant, these widening gains reflect a richer combinatorial space for division of labor rather than higher task density. Each added agent expands the opportunities for specialized roles to emerge dynamically, allowing range-limited but well-sensed robots to offload their payload to more capable teammates. This demonstrates that the architectural benefits of asynchronous handoffs compound effectively at scale.

\subsection{Interaction Between Sensing and Sample Mass}
\label{subsec:exp_sensing_mass}

We evaluate 120 base instances at all 25 combinations of five sample masses and five sensor-noise ratios, yielding a total of 3,000 runs. All other properties remain fixed. Each instance has $R=3$ robots and a total of 18 nodes. Energy budgets use $50\%$ of the reference cost of visiting all informative nodes with weightless samples ($\lambda=0$), divided by team size, and remain fixed across both sweeps. We vary sample mass over $\lambda\in\{0.1,0.5,1,2,3\}$ and the noise-variance ratio over $\sigma^2_{\mathrm{worst}}/\sigma^2_{\mathrm{best}}\in\{1,3,5,20,100\}$. The two sensor classes satisfy $1/\sigma_{\mathrm{best}}^2+1/\sigma_{\mathrm{worst}}^2=40$, so that the total sensing precision budget are fixed across the sweep.

\begin{figure}[t] 
\centering \includegraphics[width=\columnwidth]{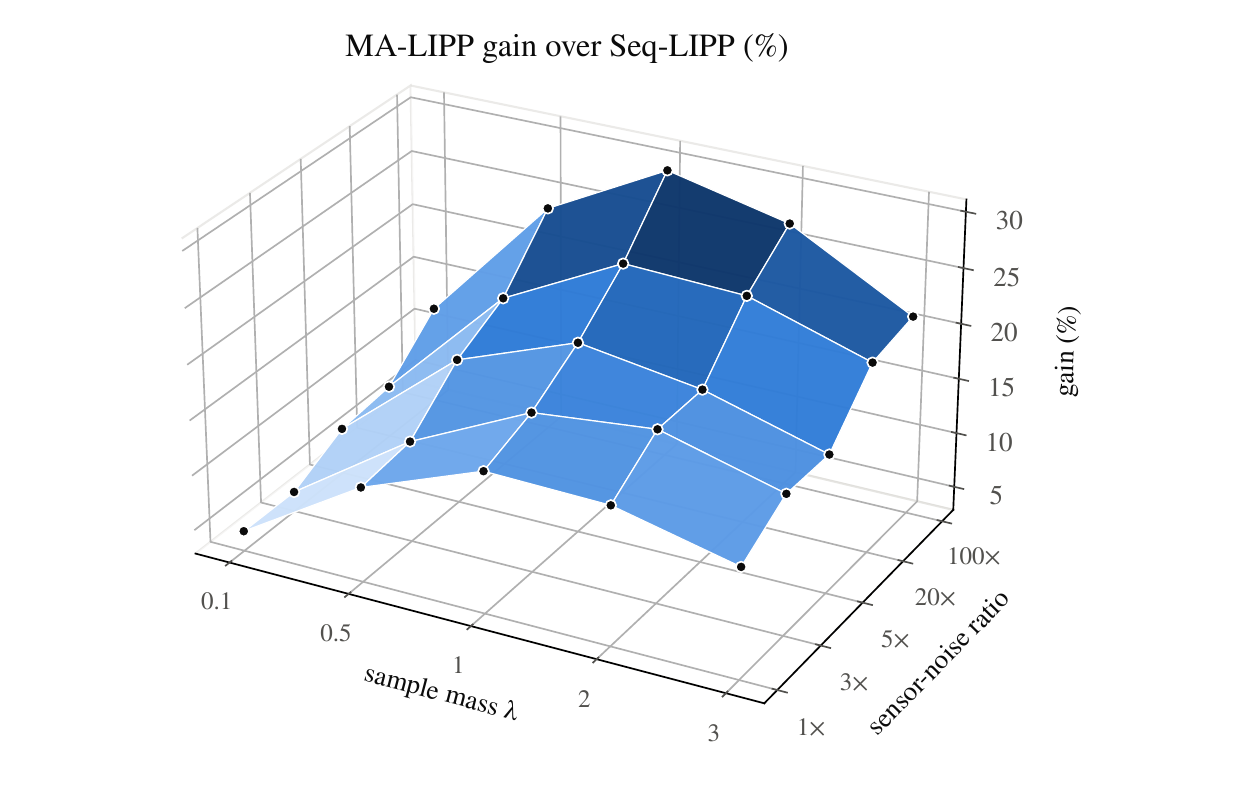} 
\caption{Mean information gain of \malipp over \seqlipp across sample mass $\lambda$ and sensor-noise ratio ($\sigma^2_{\mathrm{worst}}/\sigma^2_{\mathrm{best}}$). Surface height indicates the percentage reduction in weighted posterior variance across 120 base instances per configuration. \emph{Takeaway:} Greater sensor disparity favors cooperation, but handoffs offer limited benefit when samples are too light to make transport costly or too heavy for the available energy budget to support much collection.}
\label{fig:exp2} 
\end{figure}

Figure~\ref{fig:exp2} shows that the benefit of \malipp increases monotonically with sensor disparity at every tested sample mass, with no observed saturation up to a noise-variance ratio of $100$. Greater disparity strengthens the advantage of separating sample acquisition from delivery: a robot with a better sensor can collect samples while a more capable carrier focuses on transporting them. In contrast, the benefit peaks at intermediate sample masses. Light samples offer limited transport savings from handoffs, whereas very heavy samples restrict feasible collection and delivery, leaving fewer opportunities for cooperation. 

\subsection{Scalability and Heuristic Quality}
\label{subsec:exp_scalability}

The optimality study evaluates 334 instances, specifically those where the exact MIQP solver successfully found and certified the global optimum, spanning $R\in\{2,3,4,6,8\}$ and $n\in\{10,12,14,16,18,20\}$ for both methods. The scalability study uses 180 instances divided equally among $(R,n)\in\{(8,31),(10,38),(12,45)\}$. Map size grows with team size to keep baseline sampling coverage near $53$--$56\%$. In this study, \seqlipp receives $1200\,\mathrm{s}$ and LNS receives $1800\,\mathrm{s}$, including its sequential initialization.

\begin{figure}[t] 
\centering 
\includegraphics[width=\columnwidth]{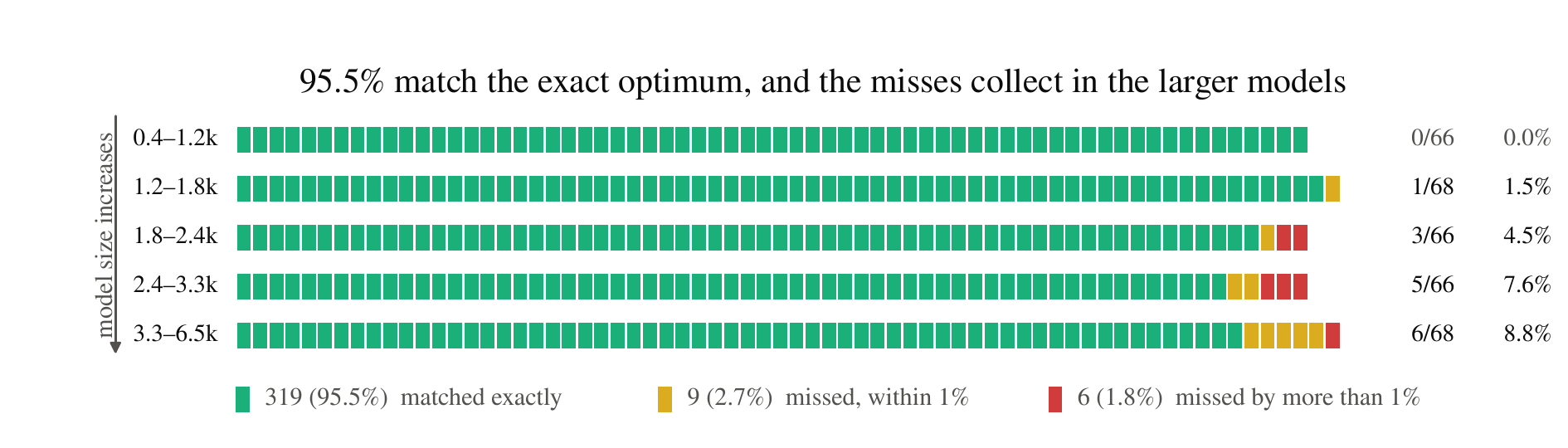}\par \vspace{0.5em} 
\includegraphics[width=\columnwidth]{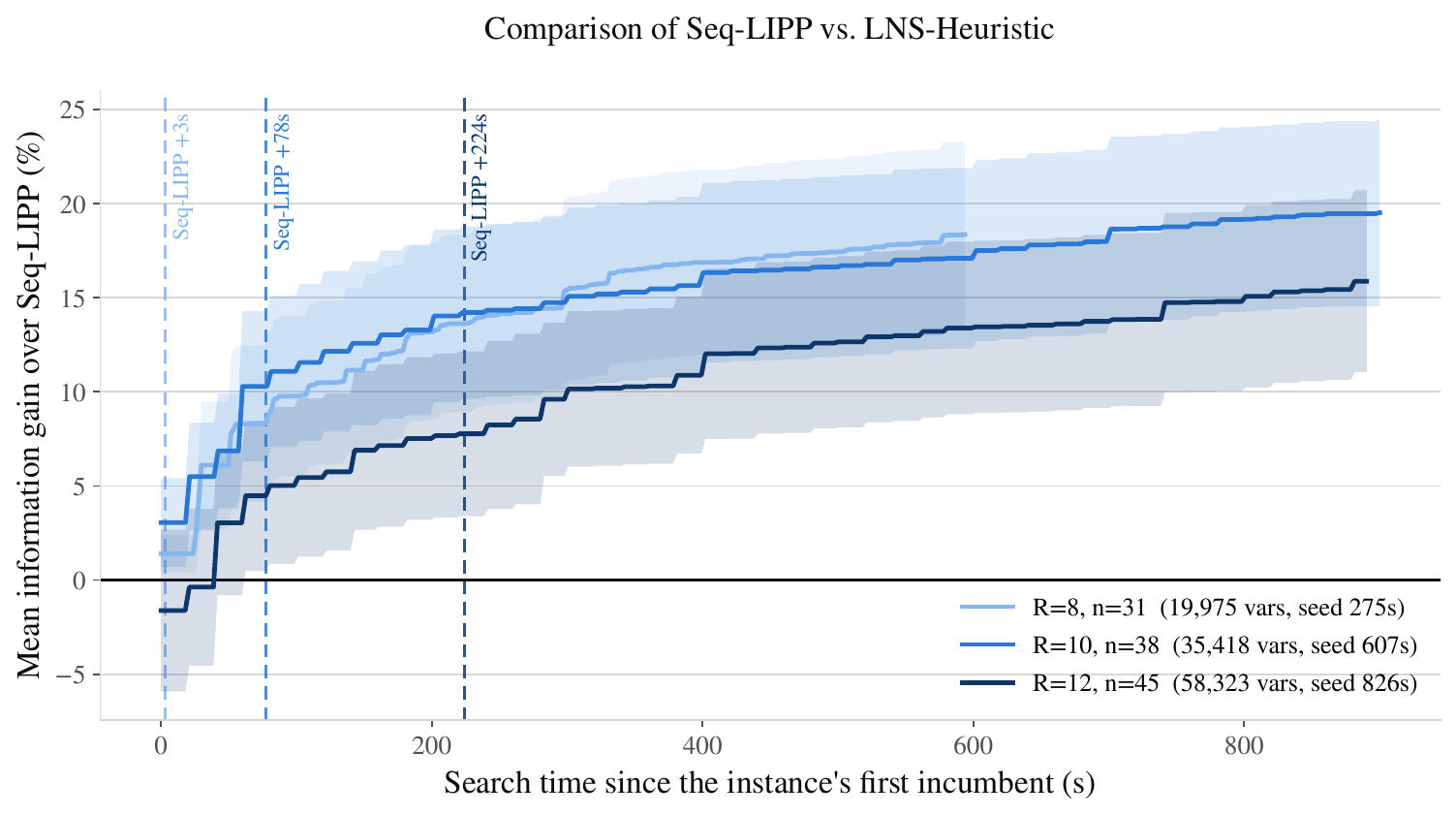} 
\caption{\textbf{Top}: Evaluation of LNS solution quality on the 334 instances where the exact MIQP solver successfully certified the global optimum, grouped by problem scale, with suboptimality rates reported at right. \textbf{Bottom}: mean posterior-variance reduction over \seqlipp versus search time since each run's first incumbent. Shading denotes $95\%$ confidence intervals; dashed lines mark mean baseline completion offsets. Each curve ends at its group's shortest search duration to retain all 60 instances; an earlier endpoint does not indicate that all runs have stopped. \emph{Takeaway:} LNS matches exact optima in $95.5\%$ of certified cases and reaches a better plan earlier than the baseline finishes, with performance steadily increasing as we provide more time.}
\label{fig:exp3} 
\end{figure}

Figure~\ref{fig:exp3} (top) plots the objective values reached by our LNS heuristic against these 334 certified MIQP optima. LNS finds feasible plans on all 334 instances and matches 319 of the known optima ($95.5\%$) within a relative-gap tolerance of $0.001\%$; another nine fall within $1\%$, bringing the total to $98.2\%$. Misses become more frequent with model size, rising from $0\%$ to $8.8\%$ across the five groups, but their magnitude does not consistently increase. Gaps above $1\%$ are consistent with the search becoming trapped in pairwise local optima, where improvement requires changing the routes of three or more robots together.

Figure~\ref{fig:exp3} (bottom) aligns each LNS run at its first feasible incumbent, after sequential initialization and fixed-route optimization. The horizontal axis measures subsequent search time, showing how pairwise updates progressively reduce posterior variance relative to the standalone \seqlipp baseline. Dashed vertical lines mark the mean baseline completion times relative to these starting points. LNS begins searching earlier than \seqlipp because its sequential initialization has a shorter time limit, although this may produce a weaker initial plan. By the end of the allotted runs, mean reductions reach $19.0\%$, $19.8\%$, and $16.1\%$ for eight, ten, and twelve robots, respectively.
\section{Conclusion and Future Work}
\label{sec:conclusion}

In this work, we introduced a cooperative load-aware informative path-planning problem for heterogeneous teams, addressing the joint optimization of routing, sampling, and physical payload transfer. By leveraging asynchronous dead drops, our MA-LIPP framework enables agents to cache physical samples for teammates without requiring a synchronous rendezvous, allowing specialized roles to emerge dynamically. Moreover, we provided a pairwise search heuristic that approaches exact optimality on smaller benchmarks while scaling efficiently to yield significant variance reductions on larger fleets. Finally, standardizing these mechanics into an open-source library provides both an immediate pipeline for field deployment and an extensible baseline for multi-robot sample collection.

For future work, \malipp opens the door to adaptive, field-deployed sampling teams. Because the LNS warm-starts from an incumbent solution, \malipp can be closed into an online loop: when in-situ measurements reveal an unexpectedly complex field or terrain delays disrupt the schedule, the team can re-plan its routes and handoffs mid-mission instead of executing a stale plan. On the hardware side, deploying \malipp on heterogeneous rover and aerial--ground teams will turn dead drops from a planning abstraction into a physical capability and validate our energy and timing models under real mission conditions.

\bibliographystyle{IEEEtran} \bibliography{references}

@inproceedings{cao2013multi, author={Cao, N. and Low, K. H. and Dolan, J. M.}, title={Multi-robot informative path planning for active sensing of spatial environmental phenomena: A tale of two algorithms}, booktitle={Proceedings of the 12th International Conference on Autonomous Agents and Multiagent Systems (AAMAS)}, year={2013}, pages={7--14}}

@article{dutta2025active, author={Dutta, S. and Wilde, N. and Smith, S. L.}, title={Informative path planning for active regression with {Gaussian} processes via sparse optimization}, journal={IEEE Transactions on Robotics}, volume={41}, pages={2184--2199}, year={2025}, doi={10.1109/TRO.2025.3548865}}

@article{gerkey2004formal, author={Gerkey, B. P. and Matari{\'c}, M. J.}, title={A formal analysis and taxonomy of task allocation in multi-robot systems}, journal={The International Journal of Robotics Research}, volume={23}, number={9}, pages={939--954}, year={2004}}

@article{korsah2013comprehensive, author={Korsah, G. A. and Stentz, A. and Dias, M. B.}, title={A comprehensive taxonomy for multi-robot task allocation}, journal={The International Journal of Robotics Research}, volume={32}, number={12}, pages={1495--1512}, year={2013}}

@inproceedings{emam2020adaptive, author={Emam, Y. and Mayya, S. and Notomista, G. and Bohannon, A. and Egerstedt, M.}, title={Adaptive task allocation for heterogeneous multi-robot teams with evolving and unknown robot capabilities}, booktitle={IEEE International Conference on Robotics and Automation (ICRA)}, year={2020}, pages={7719--7725}}

@article{schillinger2021adaptive, author={Schillinger, P. and Garc{\'i}a, S. and Makris, A. and Roditakis, K. and Logothetis, M. and Alevizos, K. and Ren, W. and Tajvar, P. and Pelliccione, P. and Argyros, A. and Kyriakopoulos, K. J. and Dimarogonas, D. V.}, title={Adaptive heterogeneous multi-robot collaboration from formal task specifications}, journal={Robotics and Autonomous Systems}, volume={145}, pages={103866}, year={2021}}

@article{notomista2022resilient, author={Notomista, G. and Mayya, S. and Emam, Y. and Kroninger, C. and Bohannon, A. and Hutchinson, S. and Egerstedt, M.}, title={A resilient and energy-aware task allocation framework for heterogeneous multirobot systems}, journal={IEEE Transactions on Robotics}, volume={38}, number={1}, pages={159--179}, year={2022}}

@misc{kim2026lipp, author={Kim, H. and Shi, G. and Sukhatme, G. S.}, title={{LIPP}: Load-aware informative path planning with physical sampling}, year={2026}, howpublished={arXiv:2603.06924}, url={https://arxiv.org/abs/2603.06924}}

@article{padberg1991, author={Padberg, M. and Sung, T.-Y.}, title={An analytical comparison of different formulations of the travelling salesman problem}, journal={Mathematical Programming}, volume={52}, pages={315--357}, year={1991}}

@article{oncan2009, author={{\"O}ncan, T. and Alt{\i}nel, {\.I}. K. and Laporte, G.}, title={A comparative analysis of several asymmetric traveling salesman problem formulations}, journal={Computers \& Operations Research}, volume={36}, number={3}, pages={637--654}, year={2009}}

@book{mood1974statistics, author={Mood, A. M. and Graybill, F. A. and Boes, D. C.}, title={Introduction to the Theory of Statistics}, edition={3rd}, publisher={McGraw-Hill}, address={New York, NY, USA}, year={1974}}

@book{taylor1997error, author={Taylor, J. R.}, title={An Introduction to Error Analysis: The Study of Uncertainties in Physical Measurements}, edition={2nd}, publisher={University Science Books}, year={1997}, note={Ch. 7}}

@article{Hollinger2014SamplingBased, author={Hollinger, G. A. and Sukhatme, G. S.}, title={Sampling-based robotic information gathering algorithms}, journal={The International Journal of Robotics Research}, volume={33}, number={9}, pages={1271--1287}, year={2014}, doi={10.1177/0278364914533443}}

@article{Das, author={Das, J. and Py, F. and Harvey, J. B. J. and Ryan, J. P. and Gellene, A. and Graham, R. and Caron, D. A. and Rajan, K. and Sukhatme, G. S.}, title={Data-driven robotic sampling for marine ecosystem monitoring}, journal={The International Journal of Robotics Research}, volume={34}, number={12}, pages={1435--1452}, year={2015}, doi={10.1177/0278364915587723}}

@article{Farley2020, author={Farley, K. A. and Williford, K. H. and Stack, K. M. and others}, title={{Mars 2020} mission overview}, journal={Space Science Reviews}, volume={216}, pages={142}, year={2020}, doi={10.1007/s11214-020-00762-y}}

@article{11284891, author={Nguyen, T. H. and Muller, E. and Rubin, M. R. and Wang, X. and Sibona, F. and McBratney, A. and Sukkarieh, S.}, title={A semi-autonomous robotic system for in situ soil sampling, analysis, and mapping in precision agriculture}, journal={IEEE Transactions on Field Robotics}, volume={3}, pages={22--39}, year={2026}}

@inproceedings{JakkalaA24IPP, author={Jakkala, K. and Akella, S.}, title={Multi-robot informative path planning from regression with sparse {Gaussian} processes}, booktitle={IEEE International Conference on Robotics and Automation (ICRA)}, year={2024}, pages={12382--12388}, doi={10.1109/ICRA57147.2024.10610484}}

@article{singh2009efficient, author={Singh, A. and Krause, A. and Guestrin, C. and Kaiser, W. J.}, title={Efficient informative sensing using multiple robots}, journal={Journal of Artificial Intelligence Research}, volume={34}, pages={707--755}, year={2009}, doi={10.1613/jair.2674}}

@inproceedings{schuster2019sampling, author={Schuster, M. J. and M{\"u}ller, M. G. and Brunner, S. G. and Lehner, H. and Lehner, P. and D{\"o}mel, A. and Vayugundla, M. and Steidle, F. and Lutz, P. and Sakagami, R. and Meyer, L. and Belder, R. and Smisek, M. and St{\"u}rzl, W. and Triebel, R. and Wedler, A.}, title={Towards heterogeneous robotic teams for collaborative scientific sampling in lunar and planetary environments}, booktitle={IROS Workshop on Informed Scientific Sampling in Large-scale Outdoor Environments}, year={2019}, url={https://elib.dlr.de/132829/}}

\end{document}